\documentclass[11pt]{article}

\usepackage[T1]{fontenc}
\usepackage[utf8]{inputenc}
\usepackage{newtxtext,newtxmath}

\usepackage[letterpaper,margin=1in]{geometry}
\usepackage{microtype}
\usepackage{graphicx}
\usepackage{booktabs}
\usepackage{multirow}
\usepackage{xcolor}
\usepackage{tabularx}
\usepackage{makecell}
\usepackage{soul}
\usepackage[numbers,sort&compress]{natbib}

\usepackage{url}

\usepackage[colorlinks=true,linkcolor=blue,citecolor=blue,urlcolor=blue]{hyperref}
\hypersetup{
    pdftitle={The AI-Enabled Scientific Frontier},
    pdfauthor={Gabriel Manso, Emma Fu, Neil Thompson},
    pdfsubject={Artificial intelligence and scientific analysis},
    pdfkeywords={artificial intelligence, scientific computing, statistics, computational cost, scientific discovery}
}

\renewenvironment{abstract}
    {\quotation}
    {\endquotation}

\date{}

\makeatletter
\renewcommand{\fnum@figure}{\textbf{Fig.~\thefigure}}
\renewcommand{\fnum@table}{\textbf{Table \thetable}}
\makeatother

\def\scititle{
The AI-Enabled Scientific Frontier
}

\title{\bfseries \boldmath \scititle}

\author{
    Gabriel Manso$^{1}$,
    Emma Fu$^{1}$,
    Neil Thompson$^{1\ast}$\\
    \small$^{1}$MIT CSAIL, MIT FutureTech\\
    \small$^\ast$Corresponding author. Email: neil\_t@mit.edu\\
}

\begin{document}

\maketitle
\noindent\textbf{One Sentence Summary:} AI is reshaping scientific analysis by opening an AI-enabled scientific frontier, where gains over traditional statistics often come at higher computational cost, but advances over scientific computing increasingly come at far lower cost.

\bigskip

\begin{abstract}
As artificial intelligence's capabilities improve, it is increasingly viewed as a general scientific method. But how true are these claims? Does AI outperform all techniques, or only some, and how is this changing? To assess the claims, we assemble a corpus of 2,507 head-to-head comparisons between AI and other scientific analysis techniques across 27 scientific disciplines from papers published between 2000 and early 2025. We find a profound dichotomy. Relative to traditional statistics, AI often outperforms, but at a significantly higher computational cost. But there are also nearly a quarter of cases where AI is both more expensive and performs worse than traditional statistical techniques and this fraction has been stable for a decade. Relative to scientific computing, AI often \textit{underperforms}, but at lower computational cost. This has begun to change: since 2020, AI's performance against scientific computing has notably strengthened and it now outperforms on more than half of comparisons. These patterns suggest that AI is therefore not a universal replacement for existing methods, but rather a valuable -- and improving -- part of a new AI-enabled scientific frontier.
\end{abstract}

\newpage

\section*{Introduction}

\noindent The future of science will be shaped by the analytical methods researchers choose, and by the performance they are willing to buy with computation \citep{thompson2022morecomputing}. One view of this future is that AI is emerging as a general scientific method and may become a dominant analytical approach across science \citep{Bianchini2022general,RoyalSociety2024ScienceAgeAI,wang2023scientific}. Another holds that AI should be treated as a ``normal'' technology whose advantages are substantial but context dependent \citep{narayanan2025ainormal,narayanan2025overreliance,messeri2024illusions}. The evidence assembled here supports neither simple universality nor simple skepticism, and is thus more in line with the second view. More importantly, it shows that AI is increasingly reshaping the scientific frontier itself.

Over the past six decades, advances in scientific computing have transformed how scientists interrogate nature. Fast Fourier transforms accelerated spectral analysis by orders of magnitude \citep{cooley1965fft}, finite element methods transformed continuum mechanics \citep{clough1960fem}, and molecular dynamics enabled atomic-resolution simulations \citep{rahman1964argon}. Today, many anticipate another revolution, this time enabled by artificial intelligence. Deep neural networks and large foundation models already perform tasks such as protein structure prediction, medium-range weather forecasting, and broader Earth-system prediction at or beyond long-standing analytical pipelines, often with dramatically different computational requirements \citep{Jumper2021AlphaFoldNature,Abramson2024AlphaFold3,lam2023graphcast,Price2025GenCast,Bodnar2025Aurora, Reichstein2019EarthSystemScience}.

These successes have encouraged expansive claims that AI may function as a general-purpose scientific method rather than a domain-specific tool. At the same time, there are reasons for caution. Success in one scientific setting need not translate to another; theoretical and empirical work has highlighted no-free-lunch limits, failures under distribution shift, overfitting to narrow benchmarks, and the risk of illusory understanding \citep{wolpert1997nfl,koh2021wilds,liao2021are,messeri2024illusions}. The computational budgets required to train frontier models have also risen steeply \citep{sevilla2024rising, thompson2023computational}, suggesting that cost concerns may also restrict AI's usage.

Individual cases highlight the variety of experiences that scientific disciplines are having with AI. A climate modeler may point to neural weather systems that rival or exceed leading operational baselines at far lower runtime cost \citep{lam2023graphcast,Price2025GenCast,Bodnar2025Aurora}, whereas a researcher working with tabular biomedical or economic data may find that deep models improve accuracy only inconsistently, or only after substantial tuning and compute \citep{shwartz-ziv2022tabular,grinsztajn2022why,Makridakis2018StatisticalMLForecasting,Makridakis2023StatisticalMLDLForecasting}. A notable single-domain precedent comes from clinical prediction, where a systematic review found no overall performance benefit of neural networks over logistic regression \citep{Christodoulou2019ClinicalPredictionReview}. Recent reviews and benchmark initiatives have begun to map out scientific machine learning within particular domains \citep{Thiyagalingam2022SciMLBenchmarks}, but what is missing is a cross-domain empirical map of where AI methods sit in the broader cost--performance landscape defined by traditional statistical learning and physics-based scientific computing. Without such a map, researchers risk overgeneralizing from their own experience.

In our analysis, we focus on AI's analytic capabilities, specifically its ability to replace other techniques from statistical analysis or scientific computing. This is distinct from work examining AI systems designed to assist or automate tasks currently performed by scientists themselves \citep{Karpathy2026Autoresearch,Boiko2023Coscientist,Szymanski2023ALab}. Our focus matters because analytical techniques often set both the quality ceiling and the computational bottleneck of scientific work. If AI systematically changes that trade-off, it changes not only scientific methods, but the pace and scope of scientific discovery itself \citep{Besiroglu2024economic, Jones2025AIinRD}. 

The productive question, then, is not whether AI is universally superior. It is where AI sits relative to existing alternatives, how it performs against them, and how that position and performance are changing as AI (and the other techniques) evolve over time. To answer these questions, we construct a dataset of 2,507 direct comparisons drawn from studies published between 2000 and early 2025, each reporting predictive performance and computational cost for at least two methods evaluated on the same task and dataset. We classify methods into three families---traditional statistics, scientific computing, and AI---based on how they learn from data \citep{Breiman2001TwoCultures}, and we group 27 scientific disciplines into seven clusters based on their dominant computational methods. This structure lets us test whether AI's effectiveness depends on what it replaces.

\begin{figure}[!ht]
    \centering
    \includegraphics[width=\textwidth]{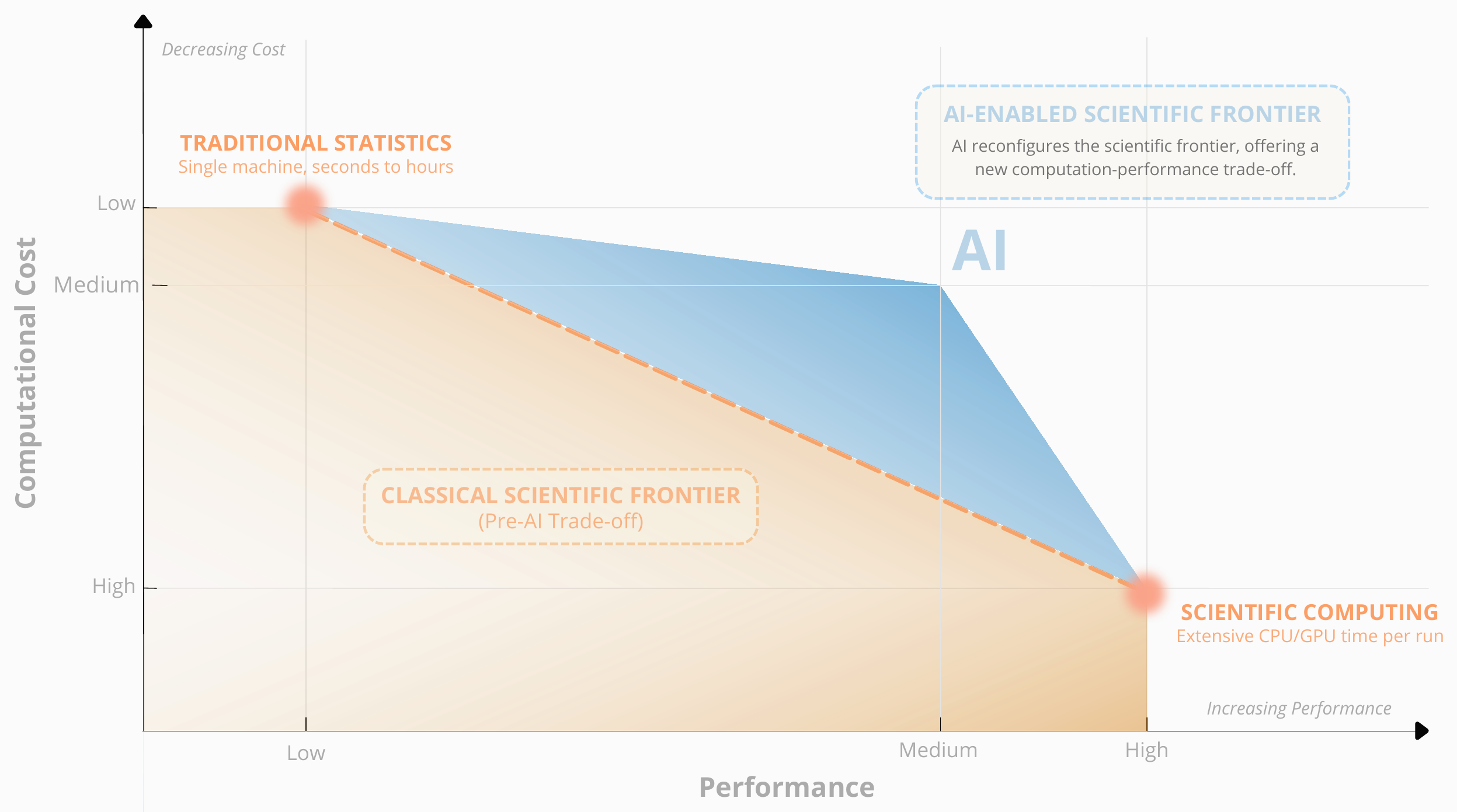}
    \caption{\textbf{The AI-enabled scientific frontier---hypothesis.} Conceptual illustration of how AI can extend the historical compute--performance trade-off defined by traditional statistics and scientific computing. In this interpretation, AI is not merely an intermediate option: it opens up a new position on the frontier that offers higher performance than traditional statistics at moderate additional cost, or much lower cost than scientific computing at a reduced level of performance.}
    \label{fig:fig1}
\end{figure}

Historically, scientific analysis has been anchored by two method families: traditional statistics, which usually offer low computational cost and solid baseline performance, and scientific computing, which offers higher-fidelity representations of complex systems but often at far greater computational expense. We hypothesize that AI can alter this landscape in two distinct ways: by outperforming traditional statistical methods at additional computational cost, or by enabling (better or worse) scientific-computing results at far lower cost. Figure~\ref{fig:fig1} illustrates this logic and introduces the reference-class structure that underlies the rest of the paper.

Five findings organize the discussion. The central one is that AI is best understood as creating a new point on the scientific frontier, rather than merely an intermediate compromise. Second, when AI replaces traditional statistical methods, the modal outcome is higher performance at higher computational cost; when AI replaces scientific-computing baselines, it provides a lower cost option that sometimes improves performance and sometimes does not. Third, the impact of AI adoption is highly domain dependent. Fourth, AI's success at replacing other methods has accelerated since 2020. And, fifth, \textbf{all} scientific methods are evolving over time, and thus AI's relative position on the scientific frontier is shifting compared to other techniques.

\subsection*{Analysis techniques across scientific domains}

\begin{figure}[!ht]
    \centering
    \vspace{-1.25cm}
    \includegraphics[width=\textwidth]{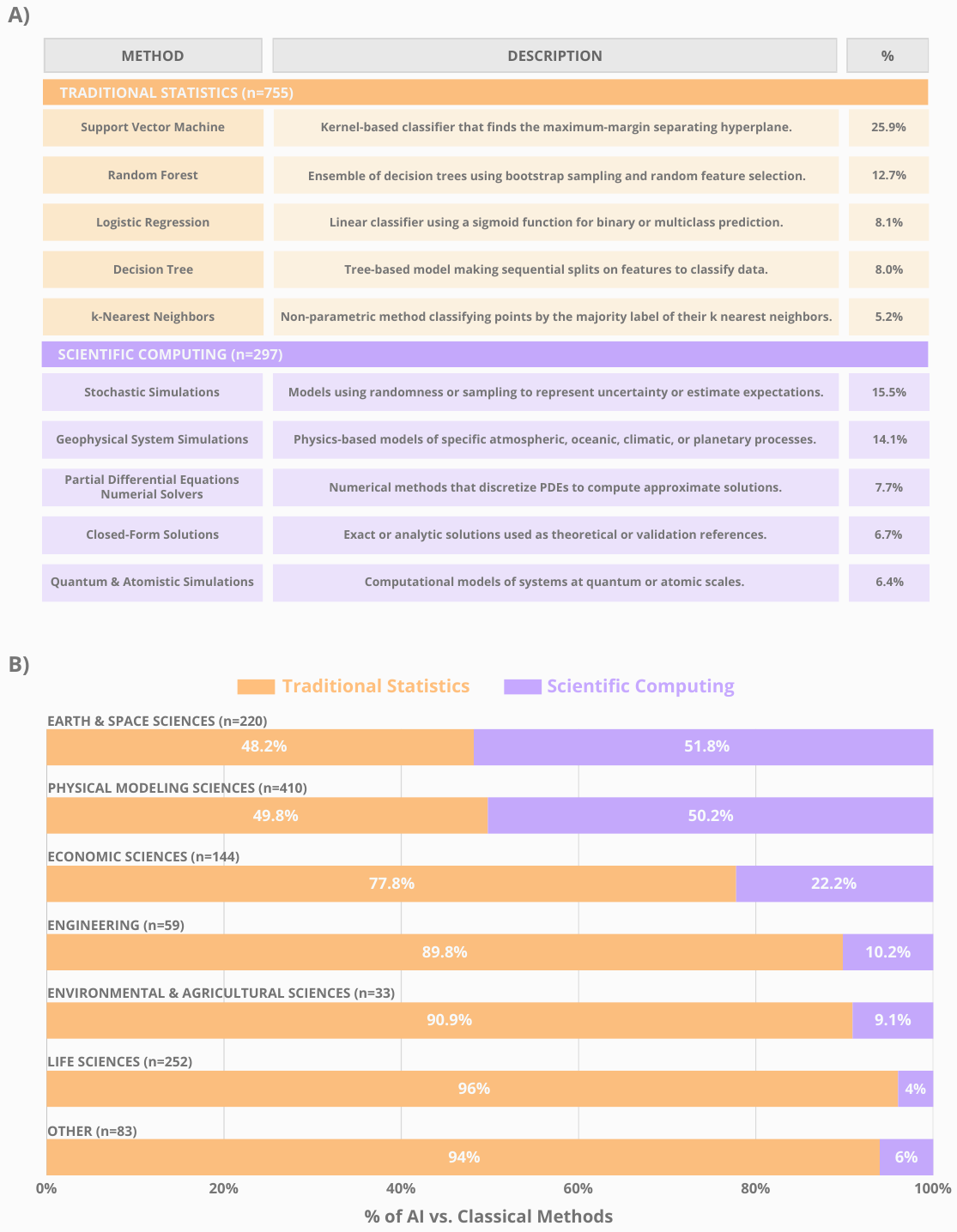}
    \caption{\textbf{AI's reference class across scientific fields.} (A) Most common traditional-statistics and scientific-computing baselines used in head-to-head AI comparisons. (B) Share of AI comparisons against each baseline family across domain clusters.}
    \label{fig:fig2}
\end{figure}

Whether AI appears transformative or incremental depends strongly on what it is being compared to. If a neural weather model is judged against an operational ensemble forecast that consumes millions of core-hours per day, even modest accuracy improvements at much lower runtime will seem consequential. If a similar neural architecture is benchmarked against a well-regularized random forest on a small tabular dataset, the relevant question is different: whether the performance gain, if any, justifies the additional computational burden. Figure~\ref{fig:fig2} makes this reference-class problem concrete. For our analysis, we infer the reference class based on the comparisons made by the literature -- that is, we consider a comparison technique to be relevant if a study incorporates and documents a comparison analysis as an integral element of its empirical or analytical framework.

On the statistical side, AI is usually compared not with trivial models but with strong workhorses such as support vector machines, random forests, logistic regression, decision trees, and $k$-nearest neighbors. On the scientific-computing side, the baselines include stochastic simulations, geophysical system models, partial differential equation solvers, analytic solutions, and quantum or atomistic simulations. These methods are often the backbone of high-fidelity scientific workflows and can be extremely computationally demanding. Figure~\ref{fig:fig2}A therefore shows that claims that ``AI performs better'' can refer to very different kinds of replacement, depending on the incumbent method family.

That heterogeneity is not incidental; it is a structural feature of the corpus. Figure~\ref{fig:fig2}B shows that in Earth \& Space Sciences and Physical Modeling Sciences, AI is evaluated almost equally often against traditional statistical and scientific-computing baselines. By contrast, Life Sciences, Engineering, Environmental \& Agricultural Sciences, and the heterogeneous ``Other'' cluster remain overwhelmingly statistics-heavy.

Figure~\ref{fig:fig2} therefore reveals two important facts. First, AI's baselines are not uniform. In some fields it is mainly trying to displace traditional statistics, whereas in others it is competing directly with mature numerical solvers. Second, where AI is evaluated against scientific computing, the baseline methods are typically high-fidelity models rather than toy simulations, implying both substantial headroom for computational savings and high standards for fidelity.

\subsection*{What happens when AI replaces existing methods?}

\begin{figure}[!htp]
    \centering
    \vspace{-1.8cm}
    \includegraphics[width=.85\textwidth]{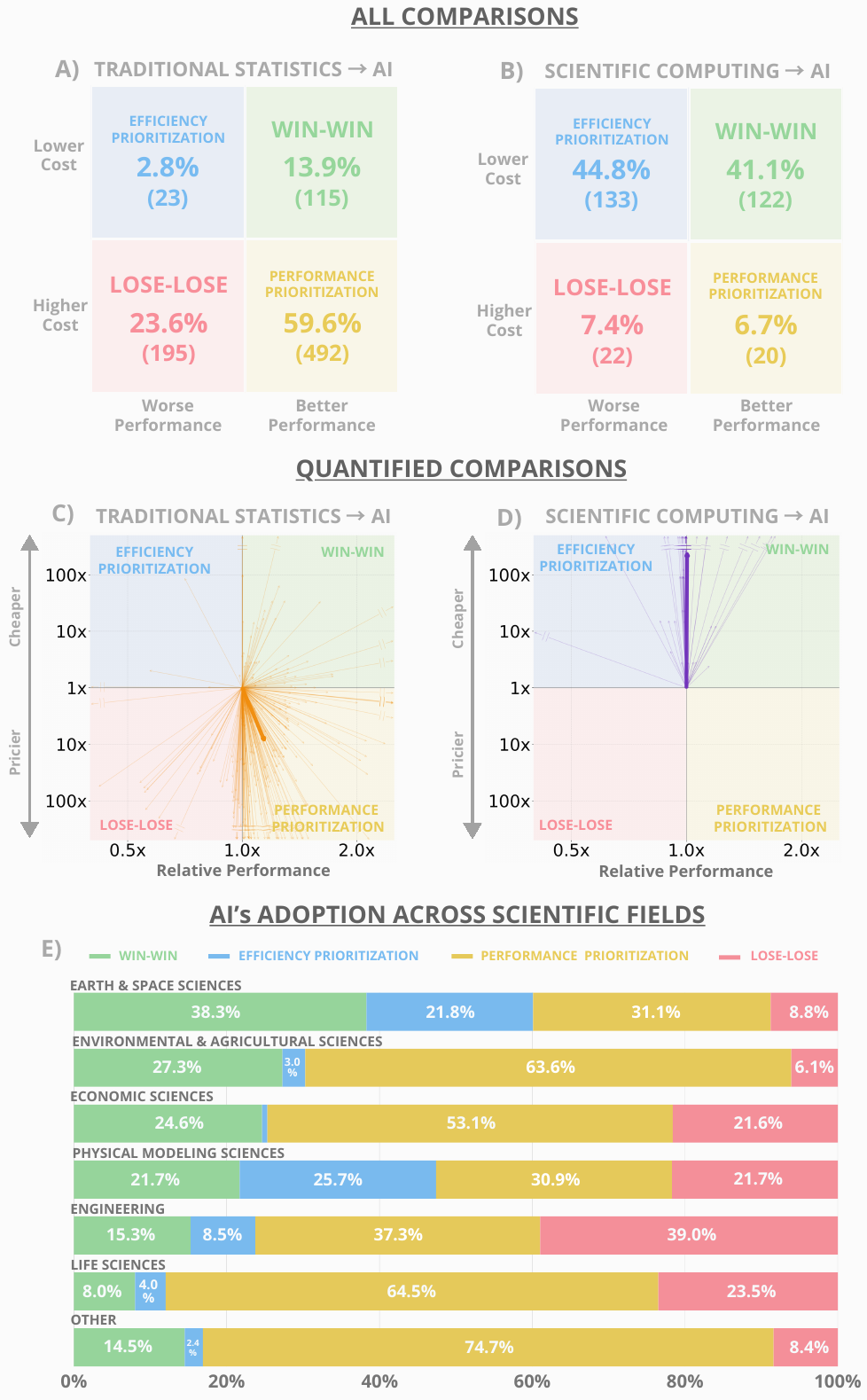}
    \caption{\textbf{AI outcomes when replacing existing methods.} (A,B) Outcome categories based on whether AI improves performance and/or computational cost relative to the baseline it replaces. (C,D) Quantitative comparisons plotted in normalized performance--cost space. (E) Outcome distributions across scientific domain clusters.}
    \label{fig:fig3}
\end{figure}

When researchers adopt AI in place of an existing method, what happens to performance and cost? Because these measures are reported in field-specific units (from accuracy and energy-per-atom to Courant-number violations and core-hours), we harmonize comparisons using two encoding schemes. With our first encoding scheme, we assign each replacement to a discrete outcome category based on whether AI improves or degrades \emph{performance} and \emph{cost}. An outcome is \textit{win-win} when AI achieves both better performance and lower cost, \textit{performance-prioritization} when AI improves accuracy but at higher cost, \textit{efficiency prioritization} when AI reduces cost but hurts performance, and \textit{lose-lose} when AI is both less accurate and more expensive than the baseline it replaces. Over all the comparisons, we find that AI achieves win-win and lose-lose outcomes in nearly equal proportions (21.1\% and 19.3\%). Whereas, AI often allows performance prioritization (45.6\%) and only sometimes efficiency prioritization (13.9\%). An even clearer pattern emerges when we disaggregate these results based on whether AI is being compared to traditional statistics or scientific computing, as shown in Figure~\ref{fig:fig3}. 

Four examples from our corpus of comparisons illustrate these scenarios:
\begin{itemize}
    \item \textbf{Win-Win: Scientific computing $\to$ AI.} Tubiana et al.\ predict protein--protein binding sites using a geometric deep learning model in place of a structural homology baseline, improving AUCPR from 0.613 to 0.694 while reducing computation time from ${\approx}30$\,days to ${\approx}1.5$\,hours \citep{Tubiana2022ScanNet}.
    \item \textbf{Performance prioritization: Traditional statistics $\to$ AI.} Davagdorj et al.\ predict patient risk of non-communicable diseases from national health survey data using a neural network in place of $k$-nearest neighbors, improving classification accuracy from 85.0\% to 95.0\% but at ${\approx}10.5\times$ longer execution time \citep{Davagdorj2021XAI}.
    \item \textbf{Efficiency prioritization: Scientific computing $\to$ AI.} George and Huerta estimate gravitational-wave parameters with a convolutional neural network (CNN) rather than a matched-filter pipeline, reducing inference time by ${\approx}10{,}000\times$ but increasing mean relative error from 18\% to 22\% \citep{George2018Multimessenger}.
    \item \textbf{Lose-Lose: Traditional statistics $\to$ AI.} Ebiwonjumi et al.\ predict the decay heat of spent light-water-reactor fuel assemblies with a neural network instead of Gaussian process regression, finding that mean absolute error rose from 4.28 to 5.56 while total training time increased by ${\approx}32\times$ \citep{Ebiwonjumi2021NuclearFuel}.
\end{itemize}

When numerical values are available, as in these examples, we compute our second encoding for outcomes: log-ratio measures that capture the magnitude of change and locate each comparison in a common cost--performance plane by anchoring the baseline technique at the origin and orienting so that positive values indicate improvement. Full details of corpus construction, taxonomy definitions, metric harmonization, and robustness checks are provided in the Supplementary Materials.

The results differ sharply depending on what AI replaces. When AI replaces traditional statistical methods (Figure~\ref{fig:fig3}A), the dominant effect is to improve performance, but at higher computational cost. The modal outcome is \emph{performance prioritization}; in 59.6\% of comparisons, AI performs better but at higher computational cost. Another 13.9\% are win-win, where AI is both better and less computationally expensive. Lose-lose cases occur  in 23.6\% of comparisons, and efficiency prioritization is rare (2.8\%). Figure~\ref{fig:fig3}C shows the quantitative distribution of these replacements, where the average performance gain is +23.2\% ($\bar{\Delta}_{\text{perf}} = 0.090$) and computational cost increases by one order of magnitude on average ($\bar{\Delta}_{\text{cost}} = 0.996$, or $\approx 10\times$), with both calculated as geometric means. This reinforces the typical trade-off in science in which better performance can be achieved at the cost of additional computation.

When AI replaces scientific computing (Figure~\ref{fig:fig3}B), the dominant effect is to reduce computational cost but about half the time this comes with a reduction in performance. More specifically, AI is win-win (better performance, lower cost) for 41.1\% of comparisons and efficiency-prioritizing for 44.8\%. Lose-lose cases are rare (7.4\%), as is performance prioritization (6.7\%). Figure~\ref{fig:fig3}D shows that when we can quantify these outcomes they are in the upper half of the plane, consistent with learned surrogates that improve performance by +26.4\% on average ($\bar{\Delta}_{\text{perf}} = 0.102$) while sharply reducing cost by approximately 2.8 orders of magnitude on average ($\bar{\Delta}_{\text{cost}} = -2.788$, or $\approx 614\times$ cheaper). These results reflect AI's ability to preserve the incumbent model's fidelity while delivering near-simulation-level performance without the associated computational burden \citep{Azizzadenesheli2024NeuralOperators}.

AI's value therefore depends on what it replaces, but not because it usually fails against traditional statistics and only succeeds against simulation. Relative to traditional statistics, AI frequently opens a performance-for-compute trade-off. Relative to scientific computing, AI is roughly split between being a cheaper-but-imperfect surrogate and being strictly better than the comparison technique.

\subsection*{Domain-specific effectiveness and the migration of effort}

Figure~\ref{fig:fig3}E shows that AI's gains are not uniform across science. Earth \& Space Sciences show the clearest broad gains, with win-win outcomes account for 38.3\% of comparisons. This is followed by Environmental \& Agricultural Sciences (27.3\%) and Economics (24.6\%). Conversely, Life Sciences has the lowest share of win-win outcomes (8.0\%) followed by Engineering (15.3\%), which is similar to the Other category (14.5\%). By contrast, lose-lose cases are most common in Engineering (39.0\%) followed by Life Sciences (23.5\%), Physical Modeling Sciences (21.7\%), and Economics Sciences (21.6\%).

A natural question is how much of this variation reflects genuine domain-specific differences in AI's effectiveness, as opposed to differences in what AI is being compared against. Because replacements of traditional statistics and of scientific computing have sharply different outcome profiles (Figures~\ref{fig:fig3}A,B), fields with more scientific-computing baselines will mechanically produce more win-win and efficiency prioritization outcomes regardless of any domain-specific effect. To disentangle the two, we weight the corpus-wide outcome rates for each baseline family by each domain's baseline mix (Figure~\ref{fig:fig2}B) to obtain expected outcome shares under the null hypothesis that AI performs identically everywhere, with variation arising solely from what it replaces. Expressed as the deviation between observed and expected rates, the domains where AI outperforms its baselines more often than composition alone would predict are Environmental \& Agricultural Sciences ($+21.0$ pp), Other ($+16.9$ pp), Earth \& Space Sciences ($+8.4$ pp), and Economic Sciences ($+4.5$ pp). Life Sciences is near parity ($+0.5$ pp), while Physical Modeling Sciences ($-9.3$ pp) and Engineering ($-17.0$ pp) fall below expectation. On the computational side, Earth \& Space Sciences ($+9.3$ pp) and Environmental \& Agricultural Sciences ($+8.6$ pp) produce cheaper-than-expected AI models, whereas Life Sciences ($-5.9$ pp) and the remaining fields show modest negative deviations.

As these deviations show, several domains deviate substantially from the null. The cleanest test comes from Earth \& Space Sciences and Physical Modeling Sciences, which share nearly identical baseline compositions (${\approx}55\%$ traditional statistics, ${\approx}45\%$ scientific computing) yet exhibit sharply divergent outcomes ($\chi^{2}=25.8$, $df=3$, $p=1.0\times10^{-5}$). Earth \& Space Sciences achieves a win-win rate of 38.3\%---twelve percentage points above its composition-predicted rate of 26.2\%---while Physical Modeling Sciences falls below the same expected rate at 21.7\%, with correspondingly elevated lose-lose outcomes (21.7\% observed vs.\ 16.4\% expected). Because baseline composition is effectively held constant in this comparison, the divergence overwhelmingly reflects domain-specific factors: differences in the structure of prediction tasks, the maturity of incumbent methods, or the intrinsic learnability of the underlying data. A second, independent line of evidence points in the same direction. Even when restricted to traditional-statistics-only comparisons---removing composition effects entirely---Life Sciences achieves a win-win rate of just 6.2\%, less than half the corpus-wide traditional-statistics average of 13.9\%, while its performance prioritization rate (66.9\%) exceeds the same average by seven percentage points. These patterns confirm that the heterogeneity in Figure~\ref{fig:fig3}E is not merely a compositional artifact. Some disciplines may feature problems whose underlying structure is intrinsically more amenable to neural approximation---for instance, spatiotemporal fields with strong physical regularities---while others may present data manifolds that are harder to compress into the continuous, low-dimensional representations that neural architectures favor \citep{Pope2021IntrinsicDim,Ansuini2019IntrinsicDim}.

Overall, AI's usage in most disciplines leads to performance prioritization, with higher costs but also higher performance. This is consistent with their wide-scale usage of baseline techniques from traditional statistics and the higher computational cost that is usually required to improve it.

\subsection*{How AI's advantage has changed over time}

\begin{figure}[!ht]
    \centering
    \includegraphics[width=\textwidth]{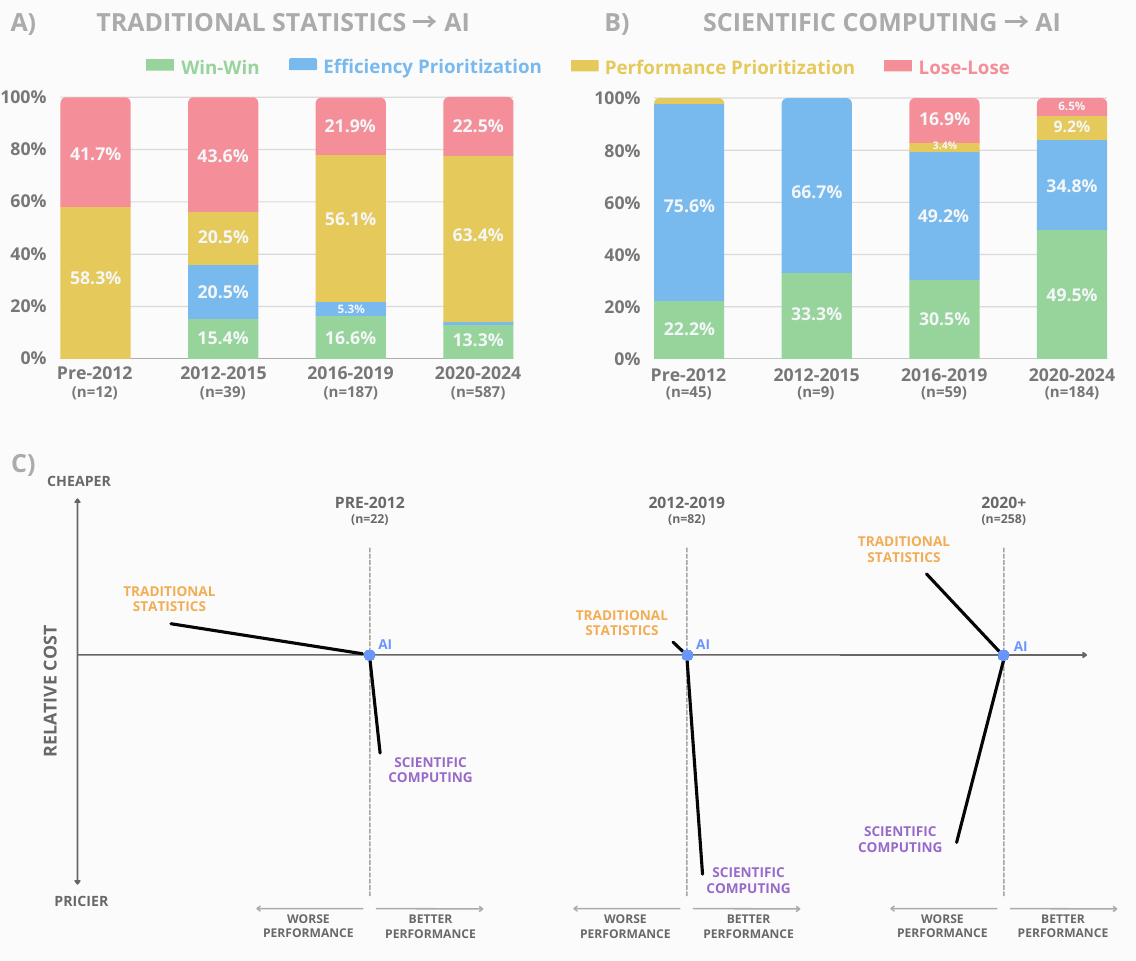}
    \caption{\textbf{How AI's advantage has changed over time.} (A,B) Outcome distributions across time when AI replaces traditional statistics and scientific computing. (C) Average normalized positions of traditional statistics, scientific computing, and AI, showing a shift from an intermediate position to AI outperforming scientific computing on average.}
    \label{fig:fig4}
\end{figure}

Figure~\ref{fig:fig4} shows that AI's relative advantage has not been static. It has changed substantially over time, and the form of that change depends on what AI is replacing.

When AI replaces traditional statistical methods (Figure~\ref{fig:fig4}A), the early evidence is limited and mixed. In the pre-2012 sample, the observed cases consist entirely of performance prioritization (58.3\%) and lose-lose (41.7\%) outcomes, indicating that early AI systems did not provide cheaper substitutions for traditional statistics. In 2012--2015, results are mixed: lose-lose outcomes account for 43.6\% of cases, while performance prioritization and efficiency prioritization each account for 20.5\%, and win-win outcomes remain limited (15.4\%). After 2015, however, the pattern seems to stabilize. In 2016--2019, performance prioritization becomes the dominant outcome (56.1\%), with lose-lose outcomes falling to 21.9\%. By 2020--2024, this pattern strengthens further: 63.4\% of substitutions are performance prioritization, whereas 22.5\% are lose-lose. The mature pattern is that AI delivers higher performance at higher computational cost \emph{and that this pattern has been stable for nearly a decade.}

The temporal pattern looks different when AI replaces scientific computing (Figure~\ref{fig:fig4}B). In the earliest period, favorable substitutions are already the norm, but they arise mainly through efficiency gains: 75.6\% of pre-2012 cases are efficiency prioritization and 22.2\% are win-win. The small 2012--2015 sample shows the same basic structure. In 2016--2019, the distribution broadens, but favorable outcomes still dominate overall: 49.2\% of cases are efficiency prioritization and 30.5\% are win-win, even though lose-lose outcomes rise to 16.9\%. By 2020--2024, the balance shifts again, this time toward stronger dominance: 49.5\% of substitutions are win-win, 34.8\% are efficiency prioritization, and lose-lose outcomes fall to just 6.5\%. Thus, relative to scientific computing, AI begins primarily as a cheaper-but-imperfect surrogate and over time some of these become strict improvements on both performance and cost.

Figure~\ref{fig:fig4}C integrates these temporal changes by locating traditional statistics, scientific computing, and AI in the normalized cost--performance plane across broad periods. Recall that in Figure~\ref{fig:fig1}, we hypothesized that AI would occupy an intermediate position on the performance--cost curve. This is indeed the pattern prior to 2020, but we can also see a new pattern emerging post 2020. More specifically, in the pre-2012 period, AI occupies an intermediate position between the two historical anchors of scientific analysis: traditional statistics are cheaper but worse-performing, whereas scientific computing is more expensive but better-performing. This same broad ordering persists in 2012--2019, but with much smaller differences in performance between the techniques. The computational cost difference in this period between AI and scientific computing is particularly pronounced. The clearest reordering appears in the 2020+ period. Traditional statistics remain cheaper but worse-performing than AI, while scientific computing becomes both more expensive and less performant on average. In this sense, the figure does not merely show AI filling the space between existing approaches. It shows AI moving from an intermediate position to one that increasingly dominates some implementations of scientific computing while maintaining a performance advantage over traditional statistics. Here, we emphasize that it is only \emph{some} implementations because even in the 2020+ period, $41\%$ of AI implementations perform worse than the scientific computing method they are compared to (and thus remain well represented by Figure~\ref{fig:fig1}).

\subsection*{The AI-enabled scientific frontier}

\begin{figure}[!ht]
    \centering
    \includegraphics[width=\textwidth]{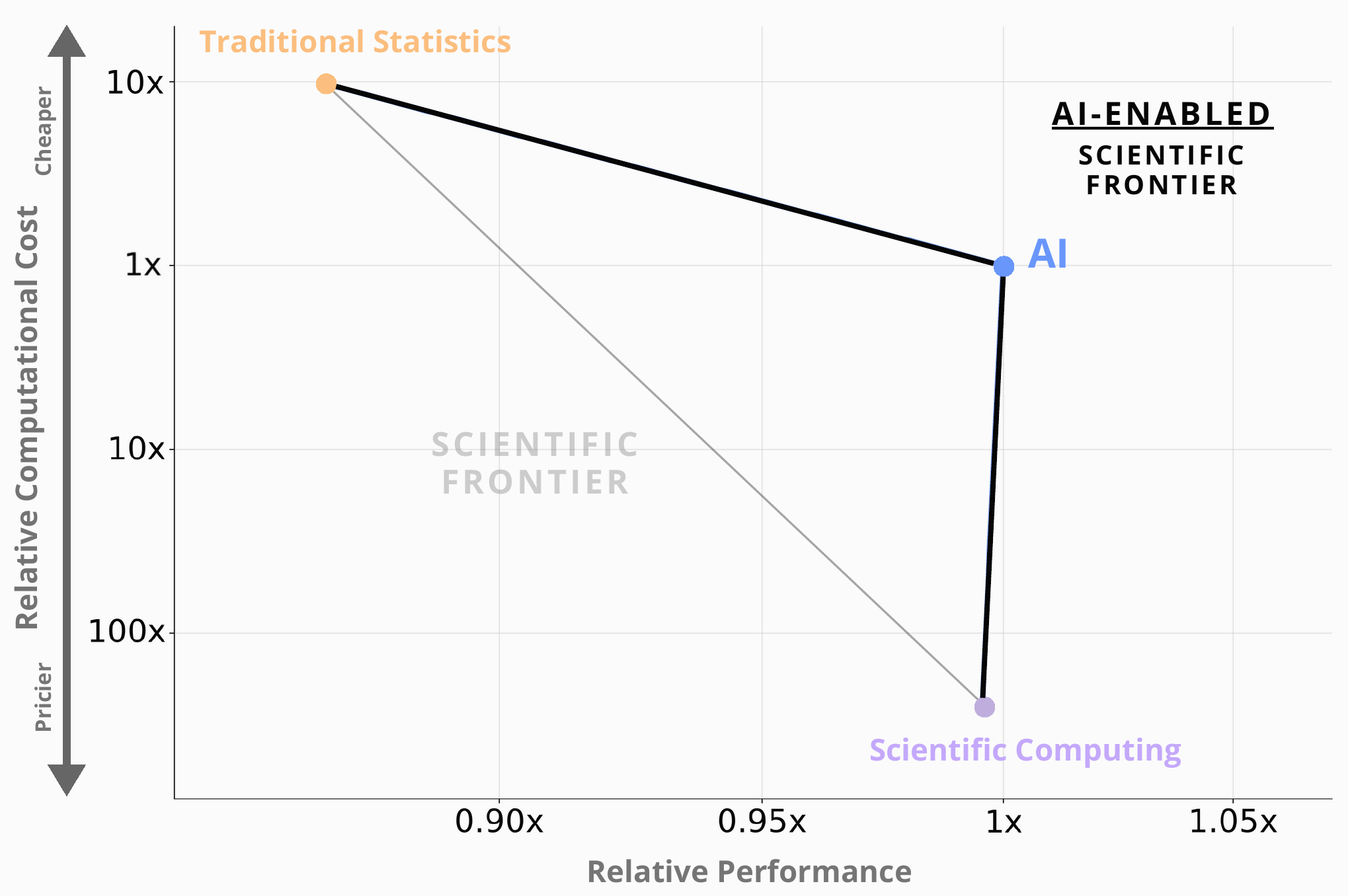}
    \caption{\textbf{The AI-enabled scientific frontier---quantified.} 
    Average positions of traditional statistics, AI, and scientific computing in normalized cost--performance space.}
    \label{fig:fig5}
\end{figure}

The previous section documented how AI's relative position has shifted over time. Figure~\ref{fig:fig5} instead takes a cross-sectional view, asking how the three method families compare when all observations are considered together in the cost--performance plane.

Traditional statistical methods occupy a low-cost, lower-performance region of the plane. Scientific computing lies at the opposite extreme, achieving higher performance but at computational costs $\approx 3{,}300\text{x}$ greater. AI occupies a distinct position. Its performance is comparable to, and on average slightly exceeds, that of scientific computing, while $\approx 330\text{x}$ cheaper. In other words, AI delivers near--simulation-level performance without simulation-level cost. Rather than simply filling the gap between traditional statistics and scientific computing, it introduces a new frontier point in the cost--performance landscape.

\begin{figure}[!ht]
    \centering
    \includegraphics[width=\textwidth]{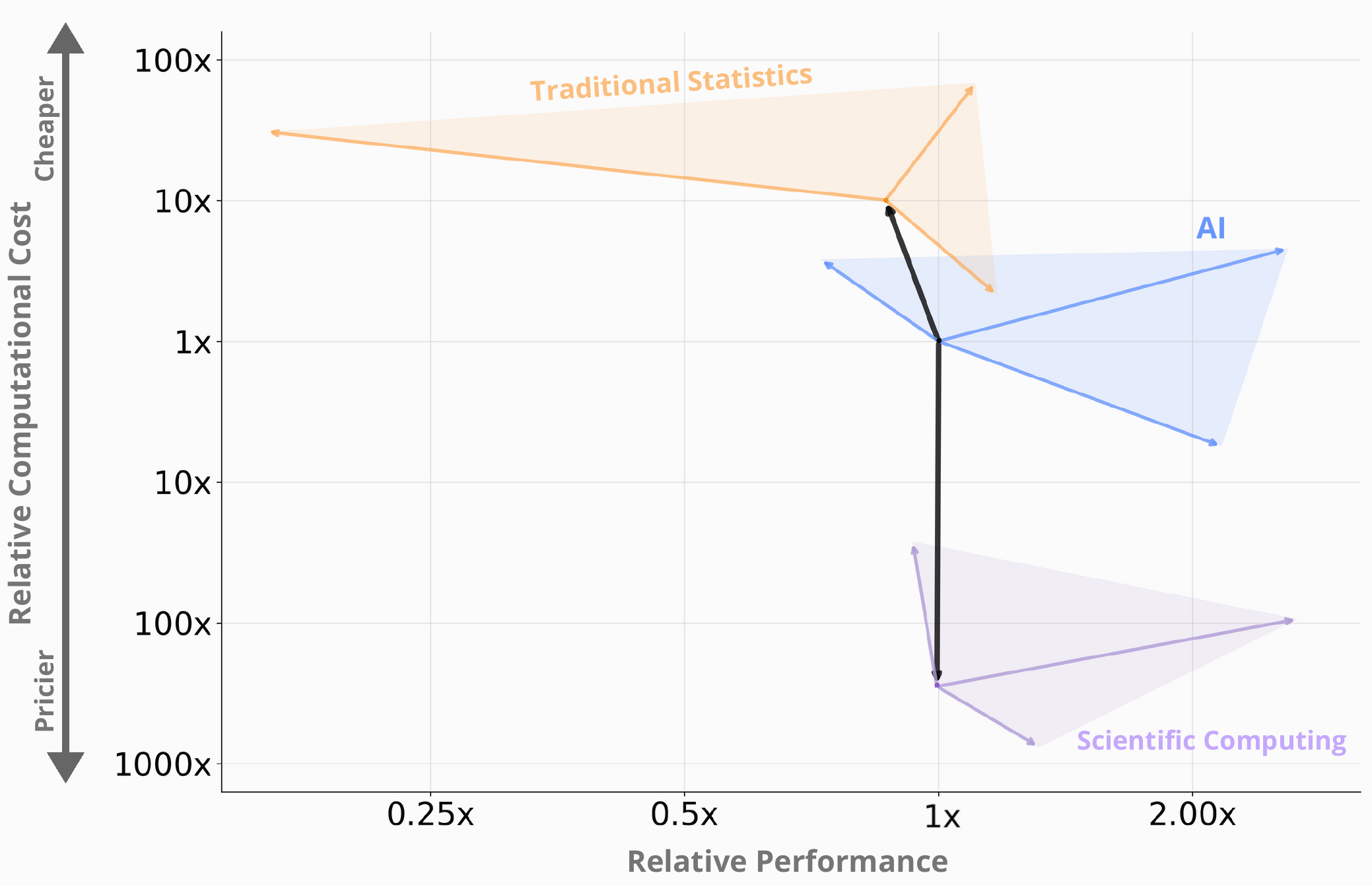}
    \caption{\textbf{Broadening regions of possibility in modern scientific methods.} 
    Distributions of observed comparisons within each method family in normalized cost--performance space.}
    \label{fig:fig6}
\end{figure}

Figure~\ref{fig:fig6} extends this perspective by adding additional head-to-head technique comparisons \textit{within} each method family (e.g.\ an older AI technique compared to a newer one). It therefore shows how each method family is evolving, with arrows indicating quadrant averages for how each family is extending the scientific frontier. The picture that emerges is of an ever-deepening set of analytical techniques that offer a broad range of cost-performance options for science.

Taken together, these results clarify how the AI-enabled scientific frontier is evolving. AI does not simply replace existing approaches, nor does it uniformly dominate across the entire cost--performance space. Instead, it expands the set of attainable trade-offs in regions where simulation-based methods have historically dominated. The resulting picture is not one of wholesale replacement but of a reconfigured frontier: traditional statistics, AI, and scientific computing occupy adjoining segments of a shared efficiency boundary, and scientific progress increasingly comes from matching problems to the segment whose inductive biases and computational profile best suit their data and accuracy requirements.

\section*{Conclusion}

Our results suggest that AI is reshaping scientific analysis, but not in a single uniform way. Relative to traditional statistical methods, AI most often improves performance by drawing on additional computation. Relative to scientific computing, AI more often preserves or improves performance while sharply reducing cost. These are distinct mechanisms. One reflects scientists' willingness to spend more computation to obtain better results; the other reflects AI's growing ability to act as a high-performing surrogate for computationally intensive pipelines.

This distinction helps explain why AI's impact varies so strongly across fields. In simulation-heavy domains, especially Earth \& Space Sciences and related physical-modeling settings, AI frequently delivers clear two-dimensional gains: lower cost with comparable or better performance. In statistics-heavy domains, the more common pattern is different. AI often improves performance, but only at higher computational cost, making its value depend on whether those gains justify the added burden of training, tuning, and deployment \citep{shwartz-ziv2022tabular,grinsztajn2022why,Makridakis2018StatisticalMLForecasting,Makridakis2023StatisticalMLDLForecasting,sculley2015hidden}. Engineering remains more mixed, underscoring that AI's scientific value depends not only on model class but also on the structure of the underlying problem.

The temporal evidence suggests that these patterns are not static. Before 2020, AI often occupied an intermediate position between traditional statistics and scientific computing. In the most recent period, AI more often exceeds scientific computing on average while maintaining a performance advantage over traditional statistics. This does not establish an inexorable trend, nor does it imply that all scientific domains will move in the same direction. It does, however, indicate that AI's role in science is becoming more expansionary than marginal.

Our study has several limitations. Only about half of the comparisons in our dataset report numerical cost metrics, and cost is measured in diverse units; we mitigate this with log-ratio encodings and robustness checks but cannot fully correct for reporting bias. The corpus largely reflects communities with strong benchmarking and model-comparison practices; fields with weaker norms around sharing baselines or reporting computational budgets are correspondingly less visible, which limits how far our conclusions can be generalized beyond the domains represented here. Our method taxonomy simplifies a continuum of architectures, and our domain clustering aggregates heterogeneous subfields. We also focus on predictive performance and computational cost, not on other dimensions of scientific value such as interpretability, calibration, robustness, or the capacity to generate novel hypotheses.

These limitations point to clear priorities for future work. First, journals and conferences could require standardized reporting of computational budgets and hardware for all methods, including baselines, enabling more systematic cost-aware evaluation \citep{Schwartz2020greenai,Henderson2020carbon}. Second, richer datasets are needed to track hybrid workflows in which AI accelerates or guides parts of numerical pipelines rather than replacing them outright. Third, predictive models linking problem characteristics, such as dimensionality, data volume, physical structure, and distribution shift, to the relative advantage of different computational approaches would help researchers and funders identify where AI is most likely to expand the scientific frontier.

Taken together, our findings support a more precise view of AI in science than either triumphalist or skeptical accounts allow. AI is neither a universal replacement for existing methods nor merely another tool within an unchanged landscape. It is a method family that increasingly opens new regions of the scientific cost--performance frontier: by purchasing performance relative to traditional statistics, and by reducing cost relative to simulation-heavy incumbents. The central question, therefore, is not whether AI will replace existing approaches everywhere, but where its inductive biases, computational profile, and surrogate capabilities allow it to create genuinely new possibilities for scientific analysis.


\section*{Acknowledgments}

We acknowledge support from MIT's UROP Program and the contributions of the master's students who assisted with early-stage data collection: Caroline C. Warren and Haley Nakamura.

\paragraph*{Funding:}

This work was supported by Open Philanthropy/Good Ventures.

\paragraph*{Data and materials availability:}
The full dataset and analysis code are available in the \href{https://github.com/MIT-FutureTech/ia-enabled-scientific-frontier}{GitHub} repository. All data needed to evaluate the conclusions in the paper are present in the paper or the Supplementary Materials.

\newpage

\bibliographystyle{sciencemag}
\bibliography{refsv2}




\newpage


\renewcommand{\thefigure}{S\arabic{figure}}
\renewcommand{\thetable}{S\arabic{table}}
\renewcommand{\theequation}{S\arabic{equation}}
\renewcommand{\thepage}{S\arabic{page}}
\setcounter{figure}{0}
\setcounter{table}{0}
\setcounter{equation}{0}
\setcounter{page}{1}


\thispagestyle{empty}

\begin{center}
\section*{Supplementary Materials for\\ \scititle}

Gabriel Manso$^{1}$,
Emma Fu$^{1}$,
Neil Thompson$^{1\ast}$\\
\small$^{1}$MIT CSAIL, MIT FutureTech\\
\small$^\ast$Corresponding author. Email: neil\_t@mit.edu\\
\end{center}

\subsubsection*{This PDF file includes:}

\noindent
\textbf{S1.} Corpus Construction and Literature Search \dotfill~\pageref{sec:search}\\
\textbf{S2.} Data Extraction and Validation \dotfill~\pageref{sec:extraction}\\
\textbf{S3.} Method Taxonomy \dotfill~\pageref{sec:taxonomy}\\
\textbf{S4.} Domain Consolidation \dotfill~\pageref{sec:domains}\\
\textbf{S5.} Metric Harmonization \dotfill~\pageref{sec:harmonization}\\
\textbf{S6.} Data Completeness \dotfill~\pageref{sec:completeness}\\
\textbf{S7.} Statistical Analyses \dotfill~\pageref{sec:stats}\\
\textbf{S8.} Sensitivity and Robustness \dotfill~\pageref{sec:sensitivity}\\[4pt]
\textbf{Tables S1--S12}

\clearpage
\renewcommand{\thepage}{S\arabic{page}}
\setcounter{page}{1}
\clearpage

\section{Corpus Construction and Literature Search}\label{sec:search}

We assembled a corpus of published studies that reported direct same-task comparisons between computational methods in scientific applications. The central inclusion principle was comparability. Each retained study had to evaluate at least two methods on a shared problem, dataset, or experimental setting, thereby permitting a meaningful assessment of relative performance and computational cost. By restricting the corpus to explicit head-to-head evaluations within the same task environment, we isolate the empirical contexts in which AI is not merely proposed, but directly tested against incumbent computational approaches.

\subsection{Databases and Sources}

Candidate papers were identified through systematic queries of five databases: IEEE Xplore, ACM Digital Library, arXiv, Google Scholar, and Scopus. Each query combined three components: domain anchor terms, AI method terms, and comparison terms. For each of the 27 disciplines, two to five query variants were constructed to capture differences in field-specific vocabulary while preserving a common search logic. Representative examples are shown in Table~\ref{tab:queries}.

\begin{table}[!ht]
\centering
\caption{Representative search queries used during corpus construction.}
\label{tab:queries}
\small
\setlength{\tabcolsep}{6pt}
\renewcommand{\arraystretch}{1.25}
\begin{tabularx}{\textwidth}{p{4cm}X}
\toprule
\textbf{Domain} & \textbf{Representative Query} \\
\midrule
Fluid Dynamics &
(``CFD'' OR ``turbulence modeling'') AND
(``neural network'' OR ``surrogate model'' OR ``deep learning'') AND
(``benchmark'' OR ``speedup'' OR ``comparison'') \\

Medicine &
(``medical imaging'' OR ``clinical prediction'') AND
(``deep learning'' OR ``convolutional neural network'' OR ``CNN'') AND
(``versus'' OR ``benchmark'' OR ``outperform'' OR ``underperform'') \\

Atmospheric Science &
(``weather prediction'' OR ``numerical weather prediction'') AND
(``neural network'' OR ``transformer'' OR ``machine learning'') AND
(``forecast skill'' OR ``computational cost'' OR ``comparison'') \\

Materials Science &
(``molecular dynamics'' OR ``density functional theory'' OR ``DFT'') AND
(``neural network'' OR ``machine learning potential'' OR ``ML potential'') AND
(``accuracy'' OR ``speedup'' OR ``benchmark'') \\

Finance &
(``financial forecasting'' OR ``stock prediction'') AND
(``deep learning'' OR ``LSTM'' OR ``neural network'') AND
(``versus'' OR ``ARIMA'' OR ``benchmark'') \\
\bottomrule
\end{tabularx}
\end{table}

\subsection{Inclusion and Exclusion Criteria}

A paper entered the screened corpus ($N = 880$) if it satisfied two core criteria. First, it had to present at least one explicit comparison between two or more computational methods on the same task. Second, the paper had to address a scientific or scientific-computing task and be available through a peer-reviewed venue or recognized preprint server.

Papers were promoted to the final dataset ($N = 268$ papers; $N = 2{,}507$ comparisons) if they additionally reported at least one performance metric for both methods and at least one computational cost indicator, whether quantitative or qualitative.

Papers were excluded if the comparison was purely theoretical, if only a single method was evaluated, if the task was outside the scope of scientific analysis, if the reported evidence was insufficient even for directional classification, or if the paper was a review rather than an original empirical study.

\section{Data Extraction and Validation}\label{sec:extraction}

\subsection{Extraction Protocol}

The canonical unit of observation is one method pair evaluated on one dataset or experimental setting. This definition prevents metric multiplication within the same experiment while preserving meaningful variation across datasets, benchmarks, or environmental conditions. When a paper reported multiple metrics for the same comparison, the metric most standard to the relevant domain was selected as the primary performance indicator. The same pair of methods evaluated on distinct datasets or settings was treated as separate observations.

For each comparison, the extraction template recorded paper metadata, domain classification, baseline and proposed method names, method-family assignments, performance metric name and values, cost metric name and values, directional labels for performance and cost, the source of any qualitative cost judgment, and free-text notes for edge cases.

To maximize consistency, information was extracted according to a fixed evidence hierarchy. Highest priority was given to tables containing numerical values, followed by explicit numerical statements in the main text, then figures, and finally qualitative author statements used only for directional classification when numerical cost data were unavailable.

\subsection{LLM-Assisted Validation}

To reduce extraction errors at scale while avoiding automation bias, we implemented a \textit{manual-first, LLM-second} validation pipeline. All comparisons were initially identified and extracted by human reviewers. The LLM system (GPT-4) was used only as a secondary verification layer and never introduced new observations or replaced human judgments without explicit review.

The validation process proceeded in six phases. First, the model mapped the structure of each paper, identifying relevant tables, figures, and result sections. Second, it enumerated candidate method comparisons within those sections. Third, it extracted the supporting evidence associated with each comparison, including reported metrics and contextual statements. Fourth, numerical values were cross-checked against the original source to verify consistency with the recorded entries. Fifth, the model classified whether its interpretation agreed with the human extraction. Finally, any disagreements were reviewed and adjudicated by a human annotator before the comparison was retained in the dataset.

\section{Method Taxonomy}\label{sec:taxonomy}

All computational methods in the corpus were classified into one of three families, corresponding to the broad methodological categories discussed in the main text and positioned along the cost--performance frontier in Figure~1.

\begin{itemize}
    \item \textbf{Traditional statistics (TS).} Algorithms that extract patterns from empirical data without relying on neural network architectures. Representative examples include support vector machines, random forests, gradient boosting, logistic regression, $k$-nearest neighbors, Gaussian processes, and principal components methods. These approaches are often computationally lighter than modern neural models, although performance depends strongly on the fit between their inductive biases and the underlying problem structure.
    
    \item \textbf{Scientific computing (SC).} Methods whose algorithmic structure is derived from embedded mathematical, physical, or mechanistic principles. Examples include Navier--Stokes solvers, finite element methods, molecular dynamics, computational fluid dynamics simulations, and quantum chemistry calculations. These approaches are often computationally demanding but are typically grounded in explicit model structure, constraints, or convergence properties.
    
    \item \textbf{AI.} Methods whose primary representational mechanism is a neural network architecture. Examples include convolutional, recurrent, and transformer networks, graph neural networks, physics-informed neural networks, neural operators, and related generative or surrogate-model architectures. These methods typically rely on large numbers of learned parameters optimized through gradient-based training.
\end{itemize}

\subsection{Replacement vs.\ Augmentation}

Beyond the family labels, cross-family AI comparisons were further classified according to whether AI replaced an incumbent method or augmented it.

\begin{itemize}
    \item \textbf{Replacement.} The proposed AI method operates independently and is evaluated as a substitute for the incumbent method.
    \item \textbf{Augmentation.} The incumbent computational pipeline remains in place, but an AI component is introduced to improve or accelerate one part of the workflow.
\end{itemize}

This distinction is important for interpretation. Replacement cases align most directly with the main paper's central question of whether AI occupies a different position on the cost--performance frontier than the methods it displaces, whereas augmentation cases more often reflect targeted optimization within an existing pipeline.

Of the $2{,}507$ comparisons in the full dataset, $1{,}201$ are cross-family AI comparisons. Among these, $1{,}122$ (93.4\%) are classified as \textit{replacement} and $79$ (6.6\%) as \textit{augmentation}. The replacement set comprises $825$ TS$\to$AI comparisons and $297$ SC$\to$AI comparisons. The remaining observations are within-family comparisons or other non-AI cross-family substitutions. The full distribution is shown in Table~\ref{tab:comptype}.

\begin{table}[!ht]
\centering
\caption{Distribution of comparison types in the full dataset ($N = 2{,}507$).}
\label{tab:comptype}
\small
\setlength{\tabcolsep}{6pt}
\renewcommand{\arraystretch}{1.18}
\begin{tabular}{@{}lr@{}}
\toprule
\textbf{Comparison Type} & \textbf{Count (\%)} \\
\midrule
\multicolumn{2}{@{}l}{\textit{Cross-family AI replacement}} \\
\quad Traditional Statistics $\to$ AI & 825 (32.9\%) \\
\quad Scientific Computing $\to$ AI & 297 (11.8\%) \\[3pt]
\multicolumn{2}{@{}l}{\textit{Other cross-family}} \\
\quad Scientific Computing $\to$ Traditional Statistics & 149 (5.9\%) \\[3pt]
\multicolumn{2}{@{}l}{\textit{Within-family}} \\
\quad AI $\to$ AI & 661 (26.4\%) \\
\quad Traditional Statistics $\to$ Traditional Statistics & 392 (15.6\%) \\
\quad Scientific Computing $\to$ Scientific Computing & 43 (1.7\%) \\[3pt]
\multicolumn{2}{@{}l}{\textit{Augmentation (hybrid)}} \\
\quad SC $\to$ SC+AI & 65 (2.6\%) \\
\quad SC $\to$ SC+TS & 29 (1.2\%) \\
\quad SC+AI $\to$ SC+AI & 21 (0.8\%) \\
\quad SC+TS $\to$ SC+AI & 14 (0.6\%) \\
\quad SC+TS $\to$ SC+TS & 11 (0.4\%) \\
\midrule
\textbf{Total} & \textbf{2,507 (100.0\%)} \\
\bottomrule
\end{tabular}
\end{table}

\section{Domain Consolidation: 27 Disciplines to 7 Clusters}\label{sec:domains}

To support cross-domain comparison without collapsing all scientific areas into a single undifferentiated group, we consolidated the 27 fine-grained disciplines into seven broader clusters. This mapping preserved meaningful domain structure while aligning fields that share similar computational tasks, data modalities, or methodological baselines (see Table~\ref{tab:domainmap}). The resulting clustering was used throughout the descriptive analyses in the supplement and main text.

\begin{table}[!ht]
\centering
\caption{Domain consolidation into seven clusters with representative characteristics and observation counts.}
\label{tab:domainmap}
\small
\setlength{\tabcolsep}{6pt}
\renewcommand{\arraystretch}{1.25}
\begin{tabularx}{\textwidth}{p{3.5cm}X X r}
\toprule
\textbf{Cluster} & \textbf{Constituent Domains} & \textbf{Shared Characteristics} & \textbf{$N$} \\
\midrule
\makecell[l]{Physical Modeling\\Sciences} &
Physics, Chemistry, Materials Science, Fluid Dynamics, Nuclear Engineering, Energy Systems, Optics and Photonics &
First-principles simulations, continuous optimization, high-dimensional regression &
970 \\

\makecell[l]{Life\\Sciences} &
Biology, Bioinformatics, Healthcare, Medicine &
Biological sequences, clinical records, imaging, heterogeneous biological data &
520 \\

\makecell[l]{Earth \& Space\\Sciences} &
Astronomy, Astrophysics, Atmospheric Science, Climate Science, Earth Science, Geophysics, Hydrology &
Observational time series, remote sensing, spatiotemporal modeling, physical constraints &
328 \\

\makecell[l]{Economic\\Sciences} &
Finance, Economics &
Financial time series, nonstationarity, strategic or adversarial dynamics &
285 \\

Engineering &
Control Systems, Robotics, Industrial Engineering, Manufacturing &
Control signals, sensor streams, real-time decision systems &
166 \\

\makecell[l]{Environmental \&\\Agricultural Sciences} &
Agricultural and Food Science, Environmental Science &
Field-scale observations, ecosystem modeling, resource management problems &
78 \\

Other &
Interdisciplinary and cross-domain studies &
Heterogeneous computational problems not fitting the primary disciplinary categories &
160 \\
\midrule
\multicolumn{3}{@{}l}{\textbf{Total}} & \textbf{2,507} \\
\bottomrule
\end{tabularx}
\end{table}

\section{Metric Harmonization}\label{sec:harmonization}

The dataset spans a highly heterogeneous measurement landscape, with more than 284 distinct performance metrics and 239 distinct computational cost metrics reported across 27 scientific domains. Because these metrics differ widely in scale, interpretation, and reporting convention, direct aggregation across studies is not generally meaningful. To preserve comparability while retaining information about the direction and, when available, the magnitude of change, we employ two complementary encoding systems.

\subsection{Binary Encoding}

For every comparison, both performance and cost are assigned directional labels:

\begin{equation}
\text{Performance binary} =
\begin{cases}
+1 & \text{new method achieves better predictive performance} \\
\phantom{+}0 & \text{equivalent within a negligible margin} \\
-1 & \text{new method achieves worse predictive performance}
\end{cases}
\label{eq:perf_binary}
\end{equation}

\begin{equation}
\text{Cost binary} =
\begin{cases}
+1 & \text{new method requires greater computational resources} \\
\phantom{+}0 & \text{equivalent} \\
-1 & \text{new method requires fewer computational resources}
\end{cases}
\label{eq:cost_binary}
\end{equation}

This directional encoding covers all 2,507 comparisons. To ensure consistency, we validated all qualitative binary labels against their corresponding quantitative log-ratios. For 'lower-is-better' performance cases, where a numerical decrease denotes a performance improvement, we inverted the corresponding log-ratios to allow for direct comparison.

\subsection{Log-Ratio Encoding}

When numerical values are available, we compute oriented log-ratio encodings:

\begin{equation}
\Delta^{X:Y}_{\text{perf}} = \log_{10}\!\left(\frac{P_X}{P_Y}\right), \qquad
\Delta^{X:Y}_{\text{cost}} = \log_{10}\!\left(\frac{C_X}{C_Y}\right)
\label{eq:logratio}
\end{equation}

where $Y$ denotes the baseline method, $X$ the proposed method, $P$ a performance metric oriented so that positive values indicate improvement, and $C$ a computational cost metric oriented so that positive values indicate increased cost.

\subsection{Outcome Categories}

Each comparison is also assigned to one of four mutually exclusive quadrants defined by the joint binary performance and cost outcomes:

\begin{itemize}
    \item \textbf{Win-Win (WW).} Performance is equal or better and cost is equal or lower.
    \item \textbf{Performance Prioritization (PP).} Performance improves, but cost increases.
    \item \textbf{Efficiency Prioritization (EP).} Performance worsens, but cost decreases or remains unchanged.
    \item \textbf{Lose-Lose (LL).} Performance worsens or remains unchanged while cost increases.
\end{itemize}

Ties (binary $= 0$) are grouped with favorable outcomes where they arise: performance ties with cost reduction ($N=36$) count as Win-Win; performance ties with cost increases ($N=16$) count as Lose-Lose; joint ties ($0$/$0$; $N=2$) count as Win-Win.

\section{Data Completeness and Selective Reporting}\label{sec:completeness}

Because the corpus draws on heterogeneous literatures with different reporting conventions, the availability of quantitative information varies across comparisons. Predictive performance metrics are usually reported numerically, whereas computational cost is often omitted or described only qualitatively. Table~\ref{tab:completeness} summarizes this difference for both directional and quantitative encodings.

\begin{table}[!ht]
\centering
\caption{Data completeness by metric type ($N = 2{,}507$).}
\label{tab:completeness}
\small
\setlength{\tabcolsep}{6pt}
\renewcommand{\arraystretch}{1.18}
\begin{tabular}{@{}lrr@{}}
\toprule
\textbf{Metric Type} & \textbf{Count} & \textbf{Percentage} \\
\midrule
\multicolumn{3}{@{}l}{\textit{Binary (directional) comparisons}} \\
\quad Performance & 2,507 & 100.0\% \\
\quad Cost & 2,507 & 100.0\% \\[4pt]
\multicolumn{3}{@{}l}{\textit{Quantitative (log-ratio) comparisons}} \\
\quad Performance & 2,170 & 86.6\% \\
\quad Cost & 1,135 & 45.3\% \\
\quad Both & 995 & 39.7\% \\
\bottomrule
\end{tabular}
\end{table}

The 41.3-percentage-point gap between quantitative performance and quantitative cost reporting is substantively important for interpretation. This gap means that analyses based on quantitative cost ratios necessarily draw on a more restricted subset of the literature than the directional analysis emphasized in the main text.

\subsection{Cost Reporting Rates by Domain Cluster}

The availability of quantitative cost data also varies across scientific domains. Some areas routinely report training time, runtime, or hardware usage, whereas others provide only qualitative descriptions of computational burden. Table~\ref{tab:costmissing} summarizes the share of comparisons with numerical cost measurements within each domain cluster.

\begin{table}[!ht]
\centering
\caption{Quantitative cost reporting rates by domain cluster (full dataset, $N = 2{,}507$).}
\label{tab:costmissing}
\small
\setlength{\tabcolsep}{6pt}
\renewcommand{\arraystretch}{1.18}
\begin{tabular}{@{}lrcc@{}}
\toprule
\textbf{Cluster} & \textbf{$N_{\text{total}}$} & \textbf{$N_{\text{quant.\ cost}}$} & \textbf{Rate} \\
\midrule
Life Sciences & 520 & 271 & 52.1\% \\
Engineering & 166 & 86 & 51.8\% \\
Physical Modeling Sci. & 970 & 468 & 48.2\% \\
Environ.\ \& Agri.\ Sci. & 78 & 27 & 34.6\% \\
Earth \& Space Sci. & 328 & 108 & 32.9\% \\
Economic Sciences & 285 & 36 & 12.6\% \\
Other & 160 & 139 & 86.9\% \\
\midrule
\textbf{Overall} & \textbf{2,507} & \textbf{1,135} & \textbf{45.3\%} \\
\bottomrule
\end{tabular}
\end{table}

These rates suggest that quantitative cost analysis is more feasible in some literatures than in others. This heterogeneity does not invalidate cross-domain comparison, but it does mean that numerical cost ratios should be interpreted with particular care in underreported fields.

\subsection{Selective Cost Reporting Test}

A potential concern in compiled datasets where computational cost is not universally reported is selective reporting. In this context, authors might be more likely to report quantitative cost metrics when the proposed AI method appears computationally favorable, while omitting such values when models impose substantial computational burdens.

To assess this possibility, we tested whether the availability of quantitative cost data depends on the reported performance outcome of the proposed AI method. Using the AI replacement sample ($N = 1{,}122$), we conducted a chi-squared test of independence between AI performance outcome (superior versus inferior/equivalent) and the presence of numerical cost data.

The test yields $\chi^2 = 0.055$ ($df = 1$, $p = 0.814$). This result provides no evidence that the availability of quantitative cost measurements depends on the reported performance outcome of the proposed method. It does not eliminate all possible reporting biases, but it suggests that the quantitative cost subsample is unlikely to be systematically biased along this particular dimension.

\section{Statistical Analyses}\label{sec:stats}

The analyses in this section focus primarily on the subset of comparisons classified as \textit{AI replacement} ($N = 1{,}122$), isolating settings in which a neural network--based method is evaluated as a direct substitute for an incumbent computational approach belonging to either Traditional Statistics (TS) or Scientific Computing (SC). Restricting attention to this subset removes hybrid augmentation cases and allows the empirical structure of direct methodological substitution to be examined more cleanly. The section is primarily descriptive, with formal inferential summaries used only to characterize association structure in the compiled literature rather than to support causal claims. Throughout, we connect the statistical patterns reported here to the conceptual argument developed in the main text while recognizing that the underlying evidence comes from heterogeneous published studies rather than a controlled benchmark environment.

\subsection{Baseline Outcome Proportions}

We begin with a descriptive summary of the outcome distribution. Table~\ref{tab:ci} reports the four outcome quadrants together with 95\% Wilson score confidence intervals, which provide more reliable coverage for bounded proportions than normal approximations.

\begin{table}[!ht]
\centering
\caption{AI replacement outcome proportions ($N = 1{,}122$) with 95\% Wilson confidence intervals.}
\label{tab:ci}
\small
\setlength{\tabcolsep}{6pt}
\renewcommand{\arraystretch}{1.18}
\begin{tabular}{@{}lccccc@{}}
\toprule
\textbf{Outcome} & \textbf{$n$} & \textbf{$N$} & \textbf{Proportion} & \textbf{95\% CI lower} & \textbf{95\% CI upper} \\
\midrule
Win-Win & 237 & 1,122 & 21.1\% & 18.8\% & 23.6\% \\
Performance Prioritization & 512 & 1,122 & 45.6\% & 42.7\% & 48.6\% \\
Efficiency Prioritization & 156 & 1,122 & 13.9\% & 12.0\% & 16.1\% \\
Lose-Lose & 217 & 1,122 & 19.3\% & 17.1\% & 21.8\% \\
\bottomrule
\end{tabular}
\end{table}

Across the corpus, the most frequent outcome when AI replaces an incumbent computational method is \textit{Performance Prioritization}, in which predictive accuracy improves while computational cost increases (45.6\% [42.7\%, 48.6\%]). Strict improvements in both dimensions (\textit{Win-Win}) occur in roughly one fifth of cases (21.1\% [18.8\%, 23.6\%]). The remaining observations are divided between \textit{Efficiency Prioritization} outcomes (13.9\%) and \textit{Lose-Lose} outcomes (19.3\%).

This descriptive pattern is consistent with the main text's broader claim that AI does not occupy a single fixed position on the cost--performance plane. In the aggregate replacement sample, neural architectures most often improve predictive capability, but these gains frequently come with added computational burden rather than universal two-dimensional improvement.

For contrast, we also summarize the much smaller set of augmentation cases, in which AI components are incorporated into existing workflows rather than replacing them outright.

\begin{table}[!ht]
\centering
\caption{AI augmentation outcome proportions ($N = 79$) with 95\% Wilson confidence intervals.}
\label{tab:ci_aug}
\small
\setlength{\tabcolsep}{6pt}
\renewcommand{\arraystretch}{1.18}
\begin{tabular}{@{}lccccc@{}}
\toprule
\textbf{Outcome} & \textbf{$n$} & \textbf{$N$} & \textbf{Proportion} & \textbf{95\% CI lower} & \textbf{95\% CI upper} \\
\midrule
Win-Win & 55 & 79 & 69.6\% & 58.8\% & 78.7\% \\
Performance Prioritization & 15 & 79 & 19.0\% & 11.9\% & 29.0\% \\
Efficiency Prioritization & 6 & 79 & 7.6\% & 3.5\% & 15.6\% \\
Lose-Lose & 3 & 79 & 3.8\% & 1.3\% & 10.6\% \\
\bottomrule
\end{tabular}
\end{table}

Augmentation scenarios show a markedly different profile. Nearly seventy percent of cases fall into the Win-Win category, suggesting that AI components are often introduced to alleviate localized bottlenecks within established computational pipelines. This contrast does not imply that augmentation is universally preferable, since the sample is small and structurally different from the replacement set, but it does reinforce the distinction drawn in the main text between substituting an incumbent method and selectively enhancing one.

\subsection{Outcome Structure by Baseline Type}

A central claim of the main paper is that the empirical role of AI depends strongly on the type of baseline method it replaces. We evaluate that claim directly by stratifying the replacement sample according to whether the incumbent method belongs to Traditional Statistics or Scientific Computing.

\begin{table}[!ht]
\centering
\caption{AI replacement outcomes stratified by baseline method type, with 95\% Wilson confidence intervals.}
\label{tab:baseline_strat}
\small
\setlength{\tabcolsep}{6pt}
\renewcommand{\arraystretch}{1.18}
\begin{tabular}{@{}lcccc@{}}
\toprule
\textbf{Baseline} & \textbf{Win-Win} & \textbf{Perf.-Prior.} & \textbf{Eff.-Prior.} & \textbf{Lose-Lose} \\
\midrule
TS ($N = 825$) & 13.9\% [11.7, 16.5] & 59.6\% [56.3, 62.9] & 2.8\% [1.9, 4.1] & 23.6\% [20.9, 26.7] \\
SC ($N = 297$) & 41.1\% [35.6, 46.8] & 6.7\% [4.4, 10.2] & 44.8\% [39.2, 50.5] & 7.4\% [4.9, 11.0] \\
\bottomrule
\end{tabular}
\end{table}

Two distinct substitution regimes emerge. When AI replaces traditional statistical methods, outcomes are dominated by the Performance Prioritization quadrant (59.6\%), indicating that neural models often achieve higher predictive performance at increased computational cost. By contrast, when AI replaces scientific simulation methods, the majority of observations fall into the Win-Win (41.1\%) or Efficiency Prioritization (44.8\%) regions.

These descriptive contrasts closely match the interpretation advanced in the main text. Relative to TS baselines, AI most often behaves as a more computationally intensive predictive model. Relative to SC baselines, AI more often functions as an empirical surrogate for computationally expensive simulation pipelines. At the same time, the nontrivial presence of Lose-Lose outcomes in both groups indicates that neither regime is universal and that substitution remains application dependent.

\subsection{Cost and Performance Effect Sizes by Baseline}

The quadrant summaries above describe directional trade-offs. We next examine the magnitude of those changes using the quantitative log-ratio encodings available for the subset of comparisons with numerical performance and cost values.

\begin{table}[!ht]
\centering
\caption{Quantitative effect sizes by baseline type (AI replacement, log$_{10}$-ratio encoding).}
\label{tab:effectsizes}
\small
\setlength{\tabcolsep}{6pt}
\renewcommand{\arraystretch}{1.18}
\begin{tabular}{@{}lccccc@{}}
\toprule
\textbf{Baseline} & \textbf{$n_\text{perf}$} & \textbf{Mean $\bar{\Delta}_\text{perf}$} & \textbf{$n_\text{cost}$} & \textbf{Mean $\bar{\Delta}_\text{cost}$} & \textbf{Median $\tilde{\Delta}_\text{cost}$} \\
\midrule
SC & 176 & $+0.102$ ($+26.4\%$) & 137 & $-2.788$ & $-2.518$ \\
TS & 794 & $+0.090$ ($+23.2\%$) & 298 & $+0.996$ & $+1.000$ \\
\bottomrule
\end{tabular}
\end{table}

Against TS baselines, AI yields average performance improvements of roughly 23\%, accompanied by a median computational cost increase of approximately one order of magnitude ($10^{1.000} = 10$). Against SC baselines, the cost pattern reverses. The mean log$_{10}$ cost shift of $-2.788$ corresponds to an average acceleration factor of approximately $10^{2.788} \approx 614$, and the median shift of $-2.518$ implies a reduction of roughly $10^{2.518} \approx 330$.

These values should be interpreted with appropriate caution. Computational cost metrics differ substantially across studies and include heterogeneous quantities such as runtime, wall-clock execution, simulation time, or hardware-linked resource usage. Accordingly, the numerical translations above are best understood as indicating large order-of-magnitude differences in the published literature rather than as universal benchmarks. Even with that caveat, the quantitative effect sizes reinforce the main paper's central asymmetry between AI substituting TS and AI substituting SC.

\subsection{Outcome Variation Across Scientific Domains}

The main text also argues that AI's position on the cost--performance frontier depends on domain context. To assess whether outcome distributions vary systematically across fields, we cross-tabulate domain cluster and outcome quadrant.

\begin{table}[!ht]
\centering
\caption{Contingency table for domain cluster by outcome quadrant in the AI replacement sample ($N = 1{,}122$).}
\label{tab:contingency}
\small
\setlength{\tabcolsep}{6pt}
\renewcommand{\arraystretch}{1.18}
\begin{tabular}{@{}lccccr@{}}
\toprule
\textbf{Cluster} & \textbf{WW} & \textbf{PP} & \textbf{EP} & \textbf{LL} & \textbf{Total} \\
\midrule
Earth \& Space Sci. & 74 & 60 & 42 & 17 & 193 \\
Economic Sciences & 33 & 71 & 1 & 29 & 134 \\
Engineering & 9 & 22 & 5 & 23 & 59 \\
Environ.\ \& Agri.\ Sci. & 9 & 21 & 1 & 2 & 33 \\
Life Sciences & 20 & 162 & 10 & 59 & 251 \\
Other & 12 & 62 & 2 & 7 & 83 \\
Physical Modeling Sci. & 80 & 114 & 95 & 80 & 369 \\
\midrule
\textbf{Total} & \textbf{237} & \textbf{512} & \textbf{156} & \textbf{217} & \textbf{1,122} \\
\bottomrule
\end{tabular}
\end{table}

The null hypothesis of independence is strongly rejected by a Pearson chi-squared test ($\chi^2 = 244.96$, $df = 18$, $p = 8.63 \times 10^{-42}$). The corresponding Cram\'{e}r's $V = 0.270$ indicates a moderate association between domain cluster and outcome type. This result supports the main paper's claim that AI's empirical role differs across scientific settings rather than following a single uniform replacement pattern.

Substantively, the table suggests that simulation-heavy areas such as Earth \& Space Sciences and Physical Modeling Sciences contribute disproportionately to Win-Win and efficiency prioritization outcomes, whereas Life Sciences and Economic Sciences are more heavily concentrated in the performance prioritization quadrant. Engineering remains notably mixed, with both Performance-Prioritization and Lose-Lose outcomes substantial. These domain-level patterns are descriptive rather than causal, but they provide an important bridge between the aggregate evidence and the field-specific discussion in the main text.

\section{Sensitivity and Robustness Analyses}\label{sec:sensitivity}

The main text describes systematic differences in the cost--performance trade-offs associated with AI substitution across scientific settings. Because the dataset aggregates comparisons from heterogeneous literature, it is important to verify that the reported patterns are not driven disproportionately by domain composition or by differences in reporting convention. In this section we therefore present two general robustness checks on the AI replacement sample.

\subsection{Leave-One-Cluster-Out Reliability}

One possible concern is that a single domain cluster could dominate the aggregate results. This is particularly relevant for simulation-heavy areas such as Physical Modeling Sciences, which contain many comparisons involving computationally intensive mechanistic baselines. To assess this possibility, we recompute the outcome distribution after sequentially removing each of the seven domain clusters from the AI replacement sample.

\begin{table}[!ht]
\centering
\caption{Leave-one-cluster-out sensitivity for the AI replacement sample. Each row removes the named cluster and reports the outcome distribution for the remaining comparisons, where WW = win-win (better performance, lower cost), PP = performance-prioritization (better performance, higher cost), EP = efficiency prioritization (worse performance, lower cost), and LL = lose-lose (worse performance, higher cost). Stable proportions across rows indicate that no single cluster drives the aggregate pattern.}
\label{tab:loocv}
\small
\setlength{\tabcolsep}{6pt}
\renewcommand{\arraystretch}{1.18}
\begin{tabular}{@{}lccccc@{}}
\toprule
\textbf{Cluster Dropped} & \textbf{$N$} & \textbf{WW\%} & \textbf{PP\%} & \textbf{EP\%} & \textbf{LL\%} \\
\midrule
Earth \& Space Sci. & 929 & 17.5 & 48.7 & 12.3 & 21.5 \\
Economic Sciences & 988 & 20.6 & 44.6 & 15.7 & 19.0 \\
Engineering & 1,063 & 21.4 & 46.1 & 14.2 & 18.3 \\
Environ.\ \& Agri.\ Sci. & 1,089 & 20.9 & 45.1 & 14.2 & 19.7 \\
Life Sciences & 871 & 24.9 & 40.2 & 16.8 & 18.1 \\
Other & 1,039 & 21.7 & 43.3 & 14.8 & 20.2 \\
Physical Modeling Sci. & 753 & 20.8 & 52.9 & 8.1 & 18.2 \\
\midrule
Full sample & 1,122 & 21.1 & 45.6 & 13.9 & 19.3 \\
\bottomrule
\end{tabular}
\end{table}

Across all exclusion scenarios the relative ordering of the outcome categories remains broadly similar. Performance-Prioritization continues to represent the largest share of outcomes, while Win-Win and Lose-Lose remain intermediate and Efficiency Prioritization remains the least common category. This indicates that the aggregate patterns reported in the main text are not solely driven by any single domain cluster. At the same time, the magnitude of the proportions varies across exclusion scenarios, which underscores that domain composition still influences the precise balance of outcomes.

\subsection{Quantitative-Cost-Only Check}

A further concern is that some studies report computational cost only qualitatively. Because such reports may be less comparable than explicit numerical measurements, we repeat the quadrant analysis using only the subset of comparisons for which numerical cost values are available.

\begin{table}[!ht]
\centering
\caption{Outcome distribution restricted to comparisons with quantitative cost measurements ($N = 435$).}
\label{tab:quant_only}
\small
\setlength{\tabcolsep}{6pt}
\renewcommand{\arraystretch}{1.18}
\begin{tabular}{@{}lccc@{}}
\toprule
\textbf{Outcome} & \textbf{Full Sample} & \textbf{Quant.\ Cost Only} & \textbf{$\Delta$} \\
\midrule
Win-Win & 21.1\% & 24.4\% & $+3.3$~pp \\
Perf.-Prioritization & 45.6\% & 44.1\% & $-1.5$~pp \\
Efficiency Prioritization & 13.9\% & 20.7\% & $+6.8$~pp \\
Lose-Lose & 19.3\% & 10.8\% & $-8.5$~pp \\
\bottomrule
\end{tabular}
\end{table}

The restricted sample produces outcome proportions that remain broadly similar to those of the full dataset, although efficiency prioritization becomes somewhat more common and Lose-Lose outcomes become less frequent. Because studies reporting quantitative cost metrics often provide more detailed benchmarking overall, this subset may represent a somewhat different population of experiments. Even so, the central structure of the outcome distribution remains intact, suggesting that the main descriptive findings are not driven solely by qualitative cost reporting.

\end{document}